\documentclass{article}
\usepackage{spconf,amsmath,graphicx}

\usepackage{url}
\usepackage{booktabs}
\usepackage{multirow}
\usepackage{amsfonts}
\usepackage[misc]{ifsym}
\usepackage{setspace}
\usepackage{bm}
\usepackage{boldline}
\usepackage{booktabs}
\usepackage[colorlinks,linkcolor=red]{hyperref}
\usepackage{url}
 
\usepackage{soul}
\usepackage{color}

\usepackage{enumitem}

\setlist{nosep, leftmargin=14pt}

\usepackage{mwe} 

\title{Advanced Deep Networks for 3D Mitochondria Instance Segmentation}
\name{Mingxing Li$^{1*}$, Chang Chen$^{1*}$, Xiaoyu Liu$^{1}$, Wei Huang$^{1}$, Yueyi Zhang$^{1, 2}$, Zhiwei Xiong$^{1, 2\dag}$}
\vspace{-0.4cm}

\address{$^1$ University of Science and Technology of China \\ $^2$ Institute of Artificial Intelligence, Hefei Comprehensive National Science Center}
\begin{document}
%
\maketitle
\begin{abstract}
\renewcommand{\thefootnote}{}
\footnotetext{$^*$ Equal contribution. $^\dag$ Corresponding author: zwxiong@ustc.edu.cn}

Mitochondria instance segmentation from electron microscopy (EM) images has seen notable progress since the introduction of deep learning methods. In this paper, we propose two advanced deep networks, named Res-UNet-R and Res-UNet-H, for 3D mitochondria instance segmentation from Rat and Human samples. Specifically, we design a simple yet effective anisotropic convolution block and deploy a multi-scale training strategy, which together boost the segmentation performance. Moreover, we enhance the generalizability of the trained models on the test set by adding a denoising operation as pre-processing. In the Large-scale 3D Mitochondria Instance Segmentation Challenge at ISBI 2021, our method ranks the 1st place. Code is available at \url{https://github.com/Limingxing00/MitoEM2021-Challenge}.



\end{abstract}

\section{Introduction}
\label{sec:introduction}
As an important kind of organelle, mitochondria provide energy for cells and are of great value to the research of life science. Generally, electron microscopy (EM) images that contain recognizable mitochondria  consume a huge storage, e.g., at the scale of Terabyte \cite{motta2019dense}.
Manual instance segmentation of mitochondria from such a large amount of data is  impossible, and automatic segmentation algorithms are highly desired. As pioneer works,
Lucci et al. \cite{lucchi2011supervoxel} propose a supervoxel-based method with learned  shape features to recognize mitochondria.
Seyedhosseini et al. \cite{seyedhosseini2013segmentation} use algebraic curves and a random forest classifier to segment mitochondria. 
Due to the limited generalizability, however, these traditional methods cannot be easily adapted to large-scale datasets such as MitoEM \cite{wei2020mitoem} including both rat and human samples.


Recently, some methods based on convolutional neural networks (CNNs) have emerged for mitochondrial segmentation.
For example, Oztel et al.\cite{oztel2017mitochondria} propose a deep network to segment 2D mitochondria slices first and then integrate 3D information with median filtering in the axial dimension.
Wei et al. \cite{wei2020mitoem} summarize the CNN-based methods  into two groups, top-down methods and bottom-up methods. Representative top-down methods use Mask-RCNN \cite{he2017mask} for instance segmentation. Due to the elongated and distorted shape of mitochondria, however, it is difficult to set a proper anchor size for Mask-RCNN in this task.
The bottom-up methods usually predict a binary segmentation mask \cite{ronneberger2015u}, an affinity map \cite{lee2017superhuman}, or a binary mask  with the instance boundary \cite{chen2016dcan}. Then a post-processing algorithm is used to distinguish instances. Although notable progress has been achieved, there is still a large room for improving the performance of mitochondria instance segmentation.


\begin{figure*}[t]
	\centering
	\includegraphics[width=0.96\textwidth]{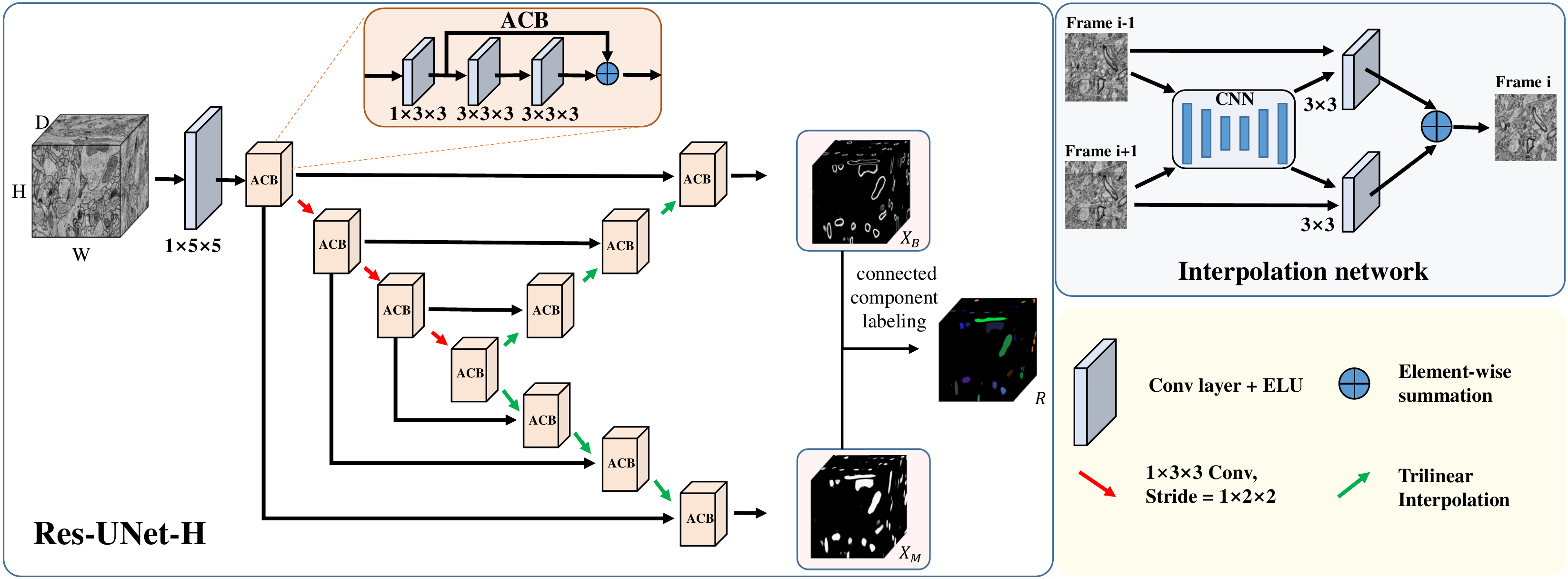} 
	\vspace{-0.8em}
	\caption{Network structure of Res-UNet-H. Note the decoder of Res-UNet-R has only one path to generate the semantic mask and the instance boundary simultaneously.}
	\label{fig:framework}
\end{figure*}

In this paper, we propose two advanced deep residual networks, named Res-UNet-R for the rat sample and Res-UNet-H for the human sample on the MitoEM dataset. Both networks generate the same form of outputs, including a semantic mask and an instance boundary. Since the human sample is more difficult (i.e., containing more noise) than the rat sample, we increase a decoder path for Res-UNet-H to predict the semantic mask and the instance boundary separately, while the decoder of Res-UNet-R has only one path. Obtaining the semantic mask and the instance boundary, we then synthesize a seed map. Finally, we adopt the connected component labeling    to obtain the mitochondria instances.

To boost the segmentation performance of our networks , we design a simple yet effective anisotropic convolution block and deploy a multi-scale training strategy.
Moreover, we observe that imaging noise is sparsely distributed on the MitoEM dataset. Especially in the human sample, the noise level is subjectively stronger than that in the rat sample. To alleviate the influence of noise on segmentation, we utilize an interpolation network \cite{niklaus2017video,huang2020learning} to restore the noisy regions coarsely marked by labor.
In addition to mitochondria instance segmentation, we also verify the proposed method has superior performance for mitochondria semantic segmentation.

\section{Method}
\label{sec:method}
\subsection{Res-UNet-R and Res-UNet-H}
We follow the bottom-up methods to extract the response map of mitochondria first. For the rat sample and the human sample, we propose two deep residual networks named Res-UNet-R and Res-UNet-H respectively. In the following description, we omit the exponential linear unit (ELU) after the convolutional layer for brevity.

\textbf{Anisotropic Convolution Block.} 
Since the MitoEM dataset has anisotropic resolution, we design an anisotropic convolution block (ACB) as shown in Fig.~\ref{fig:framework}. After a $1\times 3 \times 3$ conventional layer, we cascade two  $3\times 3 \times 3$ conventional layers to further enlarge the receptive field. At the same time, we insert the skip connection in the two  $3\times 3 \times 3$ conventional layers. 

\textbf{Network Structure.}
The overall structure of Res-UNet-H is shown in Fig.~\ref{fig:framework}. Inspired by 3D U-Net \cite{cciccek20163d}, we first embed the feature maps extracted from a 3D block with a $1\times 5 \times 5$ conventional layer. In each layer of the encoder, there is an ACB to extract the anisotropic information. Then we adopt a $1\times 3 \times 3$ conventional layer to downsample the feature maps in the lateral dimensions. In the decoder, we use the trilinear upsampling to restore the resolution of the feature maps and the ACB to reconstruct the detailed information. For Res-UNet-R, the decoder outputs a semantic mask and an instance boundary simultaneously. Since the human sample is of poorer imaging quality than the rat sample, we design two decoder paths for Res-UNet-H to predict the semantic mask and the instance boundary separately.


\textbf{Loss Function.} 
The binary cross entropy (BCE) is a common loss function used in biomedical image segmentation. To address the class imbalance problem, we adopt a weighted BCE (WBCE) loss as
\begin{equation}
L_{WBCE}(\bm{X}_i, \bm{Y}_i) = \frac{1}{DHW} \bm{W}_i L_{BCE}(\bm{X}_i, \bm{Y}_i), 
\end{equation}
where $\bm{X}_i$ and $\bm{Y}_i$ are the predicted response map and ground-truth of the i-th block, D, H, and W denote the depth, height, and width of the block, and the weight $\bm{W}_i$ is defined as
\begin{equation}
\bm{W}_i =\left\{
\begin{array}{rcl}
\bm{Y}_i + \frac{W_f}{1-W_f} (1-\bm{Y}_i)      &      & W_f > 0.5\\
\frac{1-W_f}{W_f} \bm{Y}_i + (1-\bm{Y}_i)       &      & else
\end{array} \right. 
\end{equation}
Here $W_f$ is the foreground voxel ratio, i.e., $W_f=\frac {sum(\bm{Y}_i)} {DHW}$. 
The overall loss function $L$ is defined as 
\begin{equation}
L = L_{WBCE}(\bm{X_M}, \bm{Y_M}) + L_{WBCE}(\bm{X_B}, \bm{Y_B}),
\end{equation}
where $\bm{X_M}$ and $\bm{X_B}$ are the predicted response maps of the semantic mask and the instance boundary respectively. $\bm{Y_M}$ and $\bm{Y_B}$ are the corresponding ground-truth of $\bm{X_M}$ and $\bm{X_B}$.

\subsection{Post-processing}
Obtaining the semantic mask $\bm{X_M} \in \mathbb{R}^{D\times H \times W}$ and the instance boundary $\bm{X_B} \in \mathbb{R}^{D\times H \times W}$, we can generate the seed map $\bm{S}^j ( j\in[1, D\times H \times W])$ as 
\begin{equation}
    \bm{S}^j =\left\{
    \begin{array}{rcl}
    1      &      & \bm{X_M}^j > T_1, \bm{X_B}^j < T_2 \\
    0       &      & else
    \end{array} \right. 
   \label{eq:affinity}
\end{equation}
where $T_1$ and $T_2$ are two thresholds. In our experiments, we set $T_1 = 0.9$ and $T_2 = 0.8$. Then we generate the seed map and adopt the connected component labeling   to obtain the final mitochondria instances.

\subsection{Denoising as Pre-processing}
As mentioned above, we find that by adding a denoising operation as pre-processing on the test set, the influence of noisy regions on segmentation can be alleviated, especially for the human sample.  To this end, we adopt the interpolation network initially proposed for video frame \cite{niklaus2017video} and also employed for EM image restoration in \cite{huang2020learning}. As shown in Fig.~\ref{fig:framework}, the interpolation network takes the two adjacent frames of the noisy frame as input and predicts two kernels. The two adjacent frames are then convolved by the two kernels respectively, the sum of which contributes to the restored frame. 



\section{EXPERIMENTS}
\label{sec:exp}
\subsection{Dataset}
The MitoEM dataset \cite{wei2020mitoem} consists of two (30 $\mu m$)$^3$ EM image volumes of resolution $8 \times 8 \times 30$ $nm$, which come from a rat tissue (MitoEM-R) and a human tissue (MitoEM-H) respectively. 
Each tissue has three parts, a training set ($400 \times 4096 \times 4096$), a validation set ($100 \times 4096 \times 4096$) and a test set ($500 \times 4096 \times 4096$). 
Lucchi \cite{lucchi2011supervoxel} is a mitochondria semantic segmentation dataset in which the training and test data volumns are with a size of $165 \times 1024 \times 768$.
\vspace{-0.7em}



\subsection{Implementation Details}
We adopt Pytorch (version 1.1) to implement the proposed method. Two TITAN Xp (12GB) are used for training and inference. For the MitoEM dataset, during the training stage, we adopt the data augmentation methods following \cite{wei2020mitoem}  and set the batch size as 2. The network is optimized by Adam with a fixed learning rate 0.0001. We train the network in two stages. First, we train the network in 200K iterations with the input size $32\times256\times256$  to select the best model. Then we change the input size to $36\times320\times320$ and fine-tune the network in 100K iterations. We call this two-stage training as multi-scale training (MT). 
For the Lucchi dataset, we train Res-UNet-R with only the semantic mask output, following the training details in \cite{wei2020mitoem}. Connected component labeling is no longer needed for the semantic segmentation task.


\begin{table}[!b]
\centering
\begin{tabular}{ccccc}
\hlineB{3}
Method & MitoEM-R & MitoEM-H \\ \hline
Wei \cite{wei2020mitoem} & 0.521  &0.605 \\ 
Nightingale \cite{nightingale2021automatic} & 0.715 & 0.625 \\ 
Li  \cite{li2021contrastive} &  0.870 &  0.787 \\ \hline
Ours &  \textbf{0.917} & \textbf{0.828} \\
\hline
\hlineB{3}
\end{tabular}
\caption{Instance segmentation results on the MitoEM validation set.}
\label{tab:mito}
\end{table}

\begin{table}[!b]
\centering
\begin{tabular}{ccc}
\hlineB{3}
Method & Jaccard & DSC \\ \hline
Lucchi \cite{lucchi2013learning} &0.755& 0.860\\
Liu \cite{liu2020automatic} & 0.864 & 0.926\\
Yuan \cite{yuan2020net} & 0.865 & 0.927\\
Wei \cite{wei2020mitoem}  & 0.887 & - \\
Casser \cite{casser2020fast} & 0.890 & 0.942\\
\hline
Res-UNet-R & \textbf{0.895} & \textbf{0.945} \\ 
\hlineB{3}
\end{tabular}
\caption{Semantic segmentation results on the Lucchi test set.}
\label{tab:lucchi}
\end{table}

\begin{table}[!b]
\centering
\resizebox{.97\columnwidth}{!}{
\begin{tabular}{ccccccc}
\hlineB{3}
\multirow{2}{*}{\shortstack{Instance\\boundary}} & \multirow{2}{*}{ACB}  &\multirow{2}{*}{MT}& \multicolumn{4}{c}{MitoEM-R}  \\ \cline{4-7}& & & Small  & Med  & Large  & ALL \\ 
\hline
$\times$ & $\times$ & $\times$ & 0.210 & 0.364 & 0.761  & 0.623  \\
\checkmark & $\times$ & $\times$ & 0.240     & 0.832  & 0.870    & 0.845  \\
\checkmark & \checkmark& $\times$  & \textbf{0.307}  & \textbf{0.853} & 0.935 & 0.913 \\
\checkmark & \checkmark&\checkmark& 0.277  & 0.850 & \textbf{0.949} & \textbf{0.917} \\
\hlineB{3}
\end{tabular}
}
\caption{Ablation of main components on the MitoEM-R validation set.}
\label{tab:ablation}
\end{table}

\subsection{Evaluation Metrics and Results}
We adopt an efficient 3D AP-75 metric \cite{wei2020mitoem} on the MitoEM dataset. In this case, at least 0.75 intersection over union (IoU) with the ground truth for a detection is required to be a true positive (TP). According to the number of mitochondrial voxels, mitochondria are divided into small, medium and large instances, with respective thresholds of 5K and 15K. On the Lucchi dataset, we evaluate jaccard-index coefficient (jaccard) and  dice similarity coefficient (DSC) for the foreground objects in the volumes.

As shown in Table \ref{tab:mito}, on the MitoEM validation set, the AP-75 of the proposed method surpasses the existing deep learning-based methods, Wei \cite{wei2020mitoem},  Nightingale \cite{nightingale2021automatic} and Li \cite{li2021contrastive}, by a large margin. Besides, for the semantic segmentation dataset Lucchi, the proposed method also outperforms some recent competitors, e.g., Lucchi \cite{lucchi2013learning}, Liu \cite{liu2020automatic}, Yuan \cite{yuan2020net}, Wei \cite{wei2020mitoem} and Casser \cite{casser2020fast} as shown in Table \ref{tab:lucchi}.

\subsection{Ablation Study}
\textbf{Main components.} Take Res-UNet-R as an example, the simple baseline  only generates the semantic mask and uses 3D Res block  without the MT strategy. The proposed method  adopts instance boundary and  uses 3D ACB block  with the MT strategy. We conduct an ablation study for the three main components in Table \ref{tab:ablation}, which validates their effectiveness. 
In the following subsections, we show more details about the different components and strategies.

\textbf{Block Unit Selection.} We test different block units in Table \ref{tab:rat_result} on the validation set of MitoEM-R. Here 3D SE block \cite{hu2018squeeze}, 3D ECA block \cite{wang2020eca} and 3D Res block \cite{he2016deep} are simply modified from state-of-the-art methods for the image recognition task.  In comparison with these more complex block units, our simply designed ACB alleviates overfitting and achieves the best results on the 3D mitochondria instance  segmentation task.

\textbf{Network Structure.} As shown in Table \ref{tab:human_result}, if we train and test Res-UNet-R on MitoEM-H, the AP-75 result is 0.783. By introducing an extra decoder, Res-UNet-H improves the AP-75 result to 0.816 (3.3\% increment). It verifies that Res-UNet-H can  handle more complex samples.

\textbf{Training Strategy.} As shown in Table \ref{tab:ablation} and \ref{tab:human_result}, the multi-scale training strategy (MT) we used is beneficial for both models, especially for Res-UNet-H (AP-75 improves 1.2\%). It proves that both models need larger receptive field to avoid over-fitting.

 \begin{figure}[!t]
	\centering
	\includegraphics[width=1.0\linewidth]{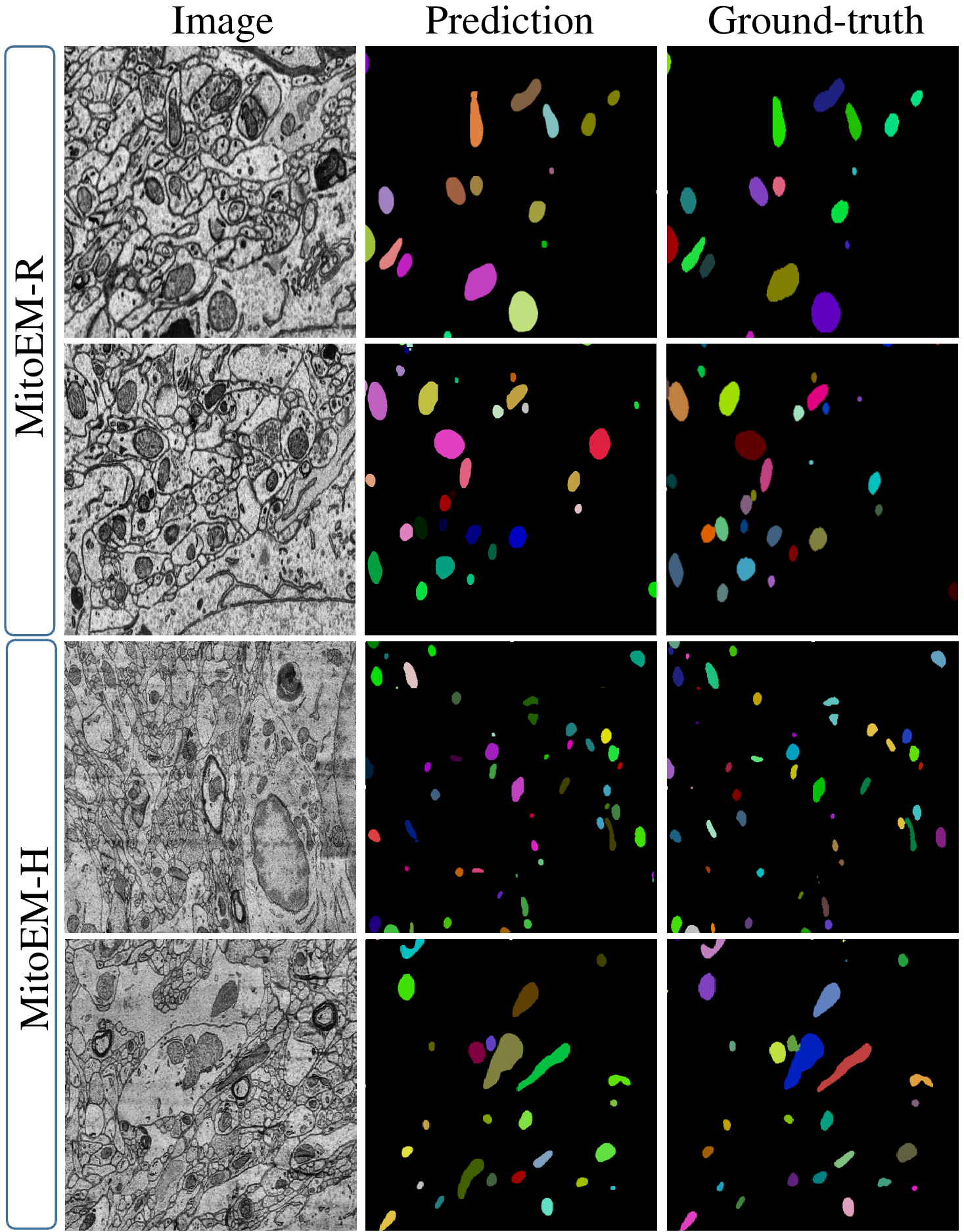}
	\vspace{-0.7cm}
	\caption{Visualized results on the MitoEM validation set.} 
	\label{fig:seg_results}
	\vspace{-0.3cm}
\end{figure}

\begin{figure}[t]
	\centering
	\includegraphics[width=1.0\linewidth]{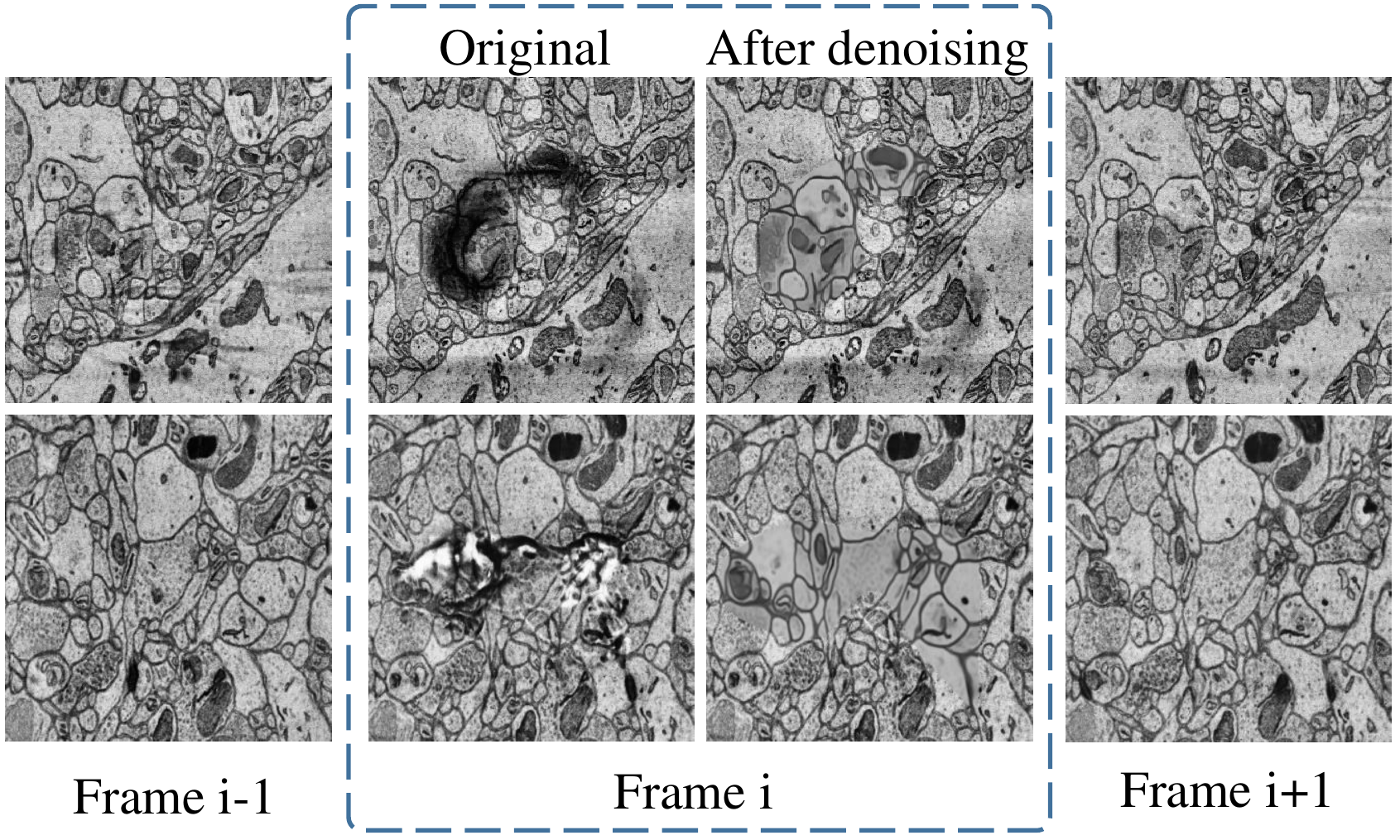} 
	\vspace{-0.7cm}
	\caption{Visualized results before and after denoising pre-processing on the MitoEM-H test set.} 
	\label{fig:Denosing}
\end{figure}

\begin{table}[!t]
\centering
\begin{tabular}{ccccc}
\hlineB{3}
\multirow{2}{*}{Block unit} & \multicolumn{4}{c}{MitoEM-R} \\ \cline{2-5} 
                 & Small  & Med  & Large  & ALL \\ \hline
2D Res block     & 0.240     & 0.832  & 0.870    & 0.845  \\
3D ECA block        & \textbf{0.398}  & 0.831 & 0.874 & 0.865 \\ 
3D SE block                & 0.388  & 0.826 & 0.891 & 0.872 \\ 
3D Res block        & 0.375  & \textbf{0.860} & 0.901 & 0.884 \\ 
\hline
3D ACB         & 0.307  & 0.853 & \textbf{0.935} & \textbf{0.913} \\
\hlineB{3}
\end{tabular}
\caption{Ablation of block unit selection on the  MitoEM-R validation set.}
\label{tab:rat_result}
\vspace{-0.3cm}
\end{table}

\begin{table}[!h]
\centering
\begin{tabular}{ccccc}
\hlineB{3}
\multirow{2}{*}{Method} & \multicolumn{4}{c}{MitoEM-H} \\ \cline{2-5} 
                 & Small  & Med  & Large  & ALL \\ \hline
Res-UNet-R          & 0.470  & 0.791 & 0.790 & 0.783 \\
Res-UNet-H          & 0.405  & 0.805 & \textbf{0.837} & 0.816 \\
Res-UNet-H+MT & \textbf{0.522} & \textbf{0.844}  & 0.826  & \textbf{0.828}  \\ 
\hlineB{3}
\end{tabular}
\caption{Ablation of network structure on the MitoEM-H validation set.}
\label{tab:human_result}
\vspace{-0.5cm}
\end{table}

\begin{table}[!h]
\centering
\begin{tabular}{ccccc}
\hlineB{3}
\multirow{2}{*}{Method} & \multicolumn{4}{c}{MitoEM-R} \\ \cline{2-5} 
                 & Small  & Med  & Large  & ALL \\ \hline
Res-UNet-R          & \textbf{0.305}   & \textbf{0.861}  & 0.848  & 0.850  \\
After denoising           & 0.151  & 0.832  & \textbf{0.854}  & \textbf{0.851}  \\
\hline
\hline
\multirow{2}{*}{Method} & \multicolumn{4}{c}{MitoEM-H} \\ \cline{2-5} 
                 & Small  & Med  & Large  & ALL \\ \hline
Res-UNet-H          & 0.522 & \textbf{0.844}  & 0.826  & 0.828  \\
After denoising          & \textbf{0.531}  & 0.834  &\textbf{ 0.827}  &  \textbf{0.829} \\
\hlineB{3}
\end{tabular}
\caption{Ablation of denoising pre-processing on the MitoEM test set.}
\label{tab:denoising test}
\end{table}

\begin{table}[!h]
\centering
\setlength{\tabcolsep}{2mm}
\begin{tabular}{cccc}
	\hlineB{3}
	Method     & MitoEM-R   & MitoEM-H   & Average        \\ \hline
	Ours       & \textbf{0.851}   & \textbf{0.829 }  &  \textbf{0.8400}  \\
	2nd    & 0.836   & 0.800   &  0.8180  \\
	3rd  & 0.833   & 0.800   &  0.8165 \\
	4th & 0.816   & 0.804   &  0.8100 \\
	\hlineB{3}
\end{tabular}
\caption{MitoEM Challenge leaderboard at the end of the testing phase.}
\vspace{-0.5cm}
\label{tab:final_rank}
\end{table}

\textbf{Denoising Pre-processing.} As shown in Fig.~\ref{fig:Denosing}, the noisy regions of the middle frame can be well restored by the interpolation network.
As shown in Table \ref{tab:denoising test}, it is demonstrated that the generalizability of the trained models can be enhanced on the test set by adding this denoising operation as pre-processing.

\vspace{-0.3cm}
\subsection{Challenge Results}
\vspace{-0.2cm}
In the Large-scale 3D Mitochondria Instance Segmentation Challenge at ISBI 2021, our method ranks the 1st place. As shown in Table \ref{tab:final_rank}, the proposed method notably outperforms other competitors on both MitoEM-R and MitoEM-H test sets.  We also show some visualized results from the validation set of MitoEM-R and MitoEM-H in Fig.\ref{fig:seg_results}. It can be seen that the predicted results by the proposed method is very close to ground-truth.

\vspace{-0.3cm}

\section{Conclusion}
\label{sec:conclusion}
In this paper, we present two advanced deep networks for 3D mitochondria instance segmentation, named Res-UNet-R for the rat sample and Res-UNet-H for the human sample.  Specifically, we exploit a simple yet effective ACB and  a multi-scale training strategy to boost the segmentation performance.  Moreover, we enhance the generalizability of the trained models on the test set by adding a denoising operation as pre-processing. 
Experimental results demonstrate the proposed method has superior performance for mitochondria instance and semantic segmentation.


\section{COMPLIANCE WITH ETHICAL STANDARDS}
Ethical approval was not required as confirmed by the license attached with the open access data (\url{https://mitoem.grand-challenge.org/}).

\section{Acknowledgement}
This work was supported in part by Anhui Provincial Natural Science Foundation Grant No. 1908085QF256 and University Synergy Innovation Program of Anhui Province No. GXXT-2019-025.

\bibliographystyle{IEEEbib}
\bibliography{strings,refs}

\end{document}